\documentclass[11pt,a4paper]{article}  
\usepackage[hyperref]{acl2018}
\usepackage{latexsym}
\usepackage{hyperref}
\usepackage{url}
\usepackage{graphicx}
\usepackage{array}
\usepackage{subfigure}
\usepackage{graphicx}
\usepackage{tablefootnote}
\usepackage{blindtext}
\usepackage[english]{babel}
\usepackage{amsmath}
\usepackage[normalem]{ulem}
\usepackage{footnote}
\usepackage{latexsym}
\usepackage{multirow}
\usepackage[linesnumbered,ruled]{algorithm2e}
\usepackage{times}
\usepackage{url}

\aclfinalcopy 

\title{A Generative Approach to Question Answering}

\author{Rajarshee Mitra \\
  Microsoft, \\
  Hyderabad \\
  {\tt ramitra@microsoft.com}  \\}

\date{}

\begin{document}
\maketitle
\begin{abstract}
Question Answering has come a long way from answer sentence selection, relational QA to reading and comprehension. We shift our attention to generative question answering (gQA) by which we facilitate machine to read passages and answer questions by learning to generate the answers. We frame the problem as a generative task where the encoder being a network that models the relationship between question and passage and encoding them to a vector thus facilitating the decoder to directly form an abstraction of the answer. Not being able to retain facts and making repetitions are common mistakes that affect the overall legibility of answers. To counter these issues, we employ copying mechanism and maintenance of coverage vector in our model respectively. Our results on MS-MARCO demonstrates it's superiority over baselines and we also show qualitative examples where we improved in terms of correctness and readability.
\end{abstract}

\section{Introduction}

Question Answering is a crucial problem in language understanding and a major milestone towards human-level machine intelligence. Datasets like SQuAD, MS-MARCO and others have led to plethora of contributions in machine reading and comprehension. The next-generation QA systems can be envisioned as the ones which can read passages and write long answers to questions. We formulate generative question answering as a form of QA where we expect the machine to produce an abstractive answer \textbf{A} that encompasses all the information from passage \textbf{P} required to answer a question \textbf{Q}. Eventually, such systems, if capable of understanding what information is necessary and what is not, will enable us to acquire information from multiple sources and present them in the form of a summarized answer.

The assumption, prevalent in most of the existing approaches, that the answer should be always a particular sub-span in the passage is a very strict one.

Generating answers should have to carefully incorporate any facts and entities which is necessary to answer the question as well as simultaneously discarding irrelevant information from P. This requires building complex relations between the question and the passage. What makes it further challenging is not only a need for good generative model but also the readability of the generations. Even if one achieves good results in terms of lexical similarity metrics like ROUGE-L, the determination of how much \textit{correctness} and \textit{readability} is preserved is a very significant concern in this task.

Our generative model is inspired by \textit{Seq2Seq} model (\citet{DBLP:conf/nips/SutskeverVL14} which is the basis of various NLG tasks like translation and summarization. In this paper, we propose our model that learns alignment between question and passage words to produce rich query-aware passage representation and using this same representation to directly decode the answer while attending to all the states in the representation. Learning end-to-end, our approach has no dependency on any external extractive labels. We also propose an approach where we make our decoder RNN state computation attention-aware by modifying the computed state of the previous step with attended encoder context.

While building such model, we noticed that, often, it replaces correct entities with similar incorrect ones (eg. correct year by incorrect year). This hinders the overall correctness of the answer being generated. To tackle this, we incorporate a \textit{copying mechanism} (\citet{DBLP:conf/acl/GuLLL16}, \citet{DBLP:journals/corr/SeeLM17}) that learns when to copy an important entity directly from the passage instead of generating anything from vocabulary. This makes our approach \textit{abstractive-cum-extractive}. Furthermore, a common error in generative models is repetitiveness in the text being generated. This also affects the generation that follows. A common practice, being used in similar tasks, like machine translation and summarization, is keeping track of a coverage vector (\citet{DBLP:conf/acl/TuLLLL16}) that keeps track of which encoder states have been attended to what extent in the past.

Our main novelty lies in demonstrating that it is possible to generate answers directly from modeled relationship between question and passage without the need to build extraction model or providing any positional labels (like start/end). We show that the decoder with the help of pointer-generator networks can itself choose to copy or generate answer words.

\section{Related Work}
Traditional QA models like \citet{DBLP:conf/icml/KumarIOIBGZPS16} and \citet{DBLP:journals/corr/SeoHF16} have shown some fascinating results in the form of relational based QA \citet{DBLP:journals/corr/WestonBCM15} or machine reading and comprehension by \citet{DBLP:journals/corr/WangJ16a}, \citep{DBLP:journals/corr/SeoKFH16} and \citet{DBLP:conf/acl/WangYWCZ17} with promising results. Their methods proved how successful is pointer networks of \citet{DBLP:conf/nips/VinyalsFJ15}. Considerable amount of work has also been done on passage ranking \citep{DBLP:journals/corr/TangDQZ17}. \citet{DBLP:journals/corr/TanWYLZ17} has taken a generative approach where they add a decoder on top of their extractive model and thus leveraging the extracted evidence to synthesize the answer. However, the model still relies strongly on the extraction to perform the generation which means it is essential to have the start and end labels (a span) for every example. Even also, when they are multiple regions in P that contribute to A, the model will need to predict multiple spans.

Our approach differs from this in the sense that it only relies on the content of Q and P to generate the answer and the extraction-abstraction \textit{soft} switch happens as a part of the learning procedure without the need for any dependency on extraction.

\section{Model Details}

Our model consists of an encoder that compute representation by using context and attentive layers and an attentive (\citet{DBLP:journals/corr/BahdanauCB14}) decoder with pointer-generator networks to decide when to copy or generate. We also keep a track of the attention at each time step to maintain a coverage vector to improve readability. 

\subsection{Representation}

We use the standard GRU (\citet{DBLP:journals/corr/ChoMGBSB14}) as the building block of our recurrence for computing representations for both questions and passages.		
            
Initially, both \textbf{P} and \textbf{Q} can be expressed by their respective word embeddings as Q = $\{w_t^Q\}_{t=1}^m$ and P = $\{w_t^P\}_{t=1}^n$.Further, the representations are built by multi-layered bi-directional GRU and weight sharing being done between \textbf{P} and \textbf{Q}.
			\begin{align}
			u_t^Q = \mathrm{BiGRU}_\theta(u_{t - 1}^Q, w_t^Q) \nonumber\\
			u_t^P = \mathrm{BiGRU}_\theta(u_{t - 1}^P, w_t^Q)
			\end{align}
         
where $u_t^Q$ and $u_t^P$ are the hidden states of the GRU at $t_{th}$ time step.

\subsection{Attentive Layer}

			\begin{figure}
				\begin{center}
					\includegraphics[width=3in]{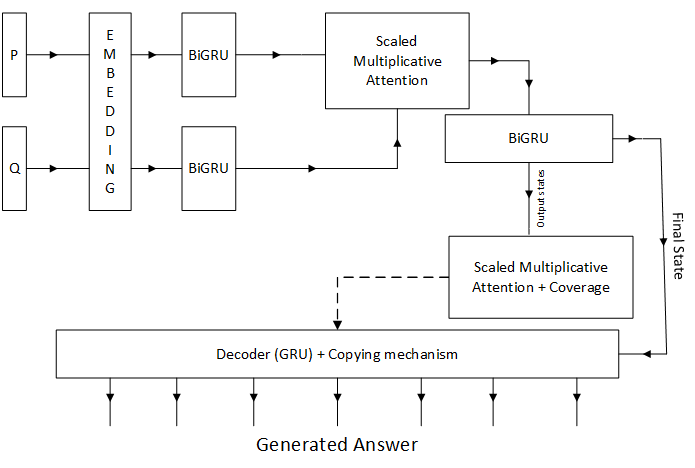}
				\end{center}
				\caption{Block diagram of our model. Both Attention1 and Attention2 is the scaled multiplicative attention we discuss in (3).}
				\label{r-network_overview}
			\end{figure}
            
We use a multiplicative attention with a scaling factor \citep{DBLP:conf/nips/VaswaniSPUJGKP17} to compute alignment between question and passage words $u_i^Q$ and $u_i^P$ respectively. Specifically, for each $u_i^P$, we take the weighted sum of all $u^Q$ which is then concatenated to $u_i^P$ to form it's final representation. We also use gating mechanism \citep{DBLP:journals/corr/TanWYLZ17} to provide varying importance to passage words (\ref{gated}). 

            			\begin{figure}
				\begin{center}
					\includegraphics[width=3.2in]{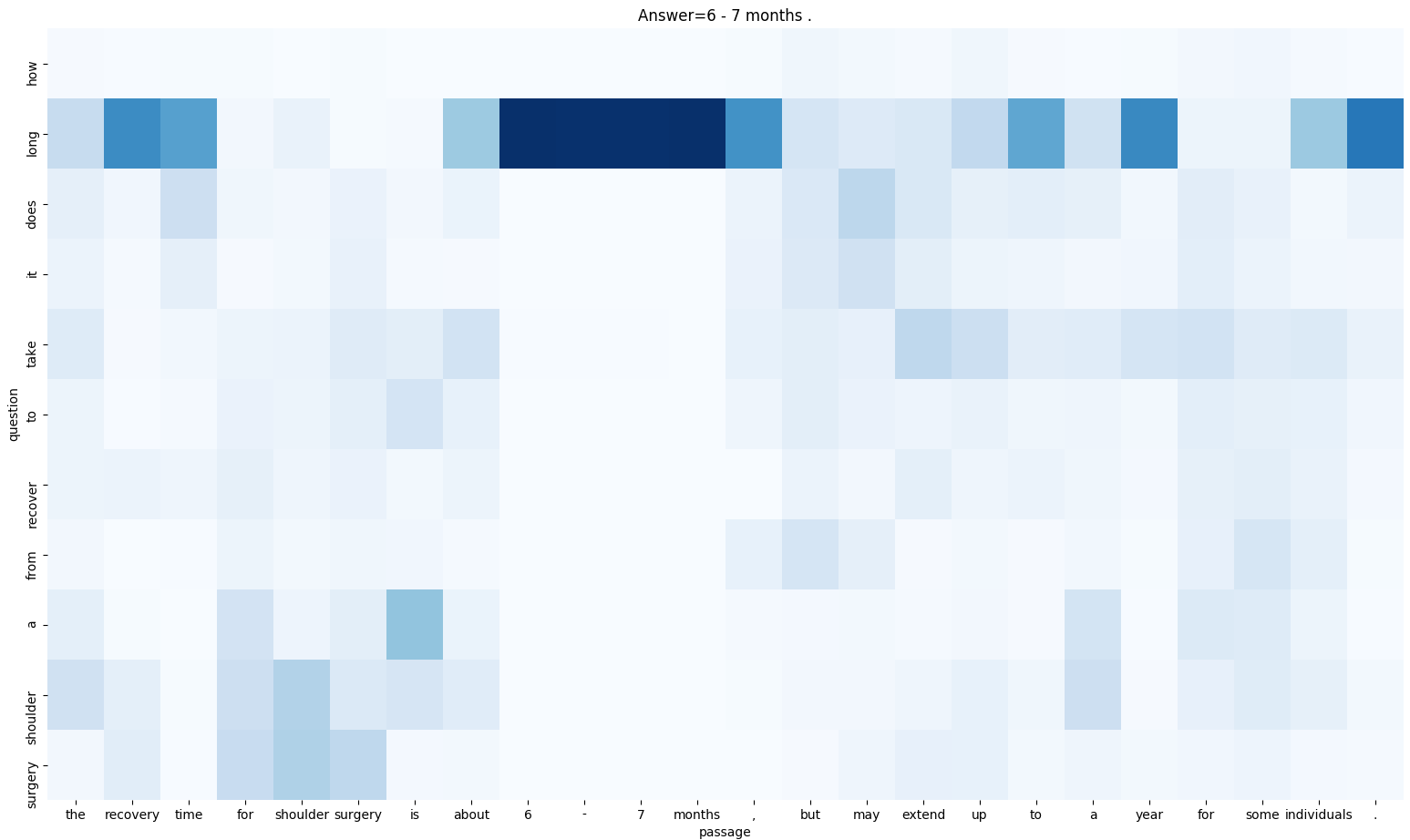}
				\end{center}
				\caption{An alignment matrix between Q and P shows how the model associates each word in Q with each word in P. As an example, when the model reads the word \textit{long} in the question, it focuses most of it's attention on \textit{6-7 months} in passage. All the columns are softmax normalized.} 
				\label{attention}
			\end{figure} 
            
Mathematically, we can express the attentive layer as:

			\begin{table*}[h]
				\centering
				\begin{tabular}{c p{13cm}}
					\hline
					\multicolumn{2}{l}{\textbf{}} \\
					& $\textbf{Question}$: when was the death penalty abolished? \\
					& $\textbf{Passage}$: The last executions UK took place in 1964 , and the death penalty was abolished in 1998 . In 2004 the UK became a party to the 13th Protocol to the European Convention on Human Rights and prohibited the restoration of the death penalty . Death penalty , capital punishment , or execution is the legal process of putting a person to death as a punishment for a crime . Modern History of Death Penalty . 1608 - Earliest death penalty in the British American Colonies handed out for UNK \\
					& $\textbf{Model with copying}$: 1998 \\
                    & $\textbf{Model without copying}$: 1964 \\
				\end{tabular}
                \caption{Replacement of correct entity by negative and similar one. We show two results from two models -- with and without point-gen}
				\label{replacement}
			\end{table*}            
            
			\begin{align}
            u^{Q^-}&=tanh(W^{Q}u^{Q}); u^{P^-}=tanh(W^{P}u^{P}) \nonumber \\
			a_{ij} &= \frac{1}{\sqrt[]{D}}(u_{i}^{P^-}.{u_j^{Q^-}}^T) [D=hidden\_dim]\label{scaled_attention} \\
            w_i &= softmax(a_i) \nonumber \\
            v_t^P&= w_t u^Q \label{attended_p} \\
            g &= \sigma(W_g[v_t^P; u_t^P]) \label{gated} \\
            [v_t^P; u_t^P]^* &= g [v_t^P; u_t^P] \nonumber \\
            c_t &= \mathrm{BiGRU}(c_{t - 1}, [v_t^P; u_t^P]^*) \label{attentive_layer}
			\end{align}

$u^{Q^-}$ and $u^{P^-}$ are non-linear ($tanh$) transformations of question and passage representations $u^Q$ and $u^P$ respectively. $a$ computes dot product attention between transformed query and passage representations followed by a scaling factor $\sqrt[]{D}$. This also implies that for each passage word, we find the weighted importance of all question words, scale them and finally a softmax normalization to produce weights. With the weights, we thus get, we find the weighted sum of all the question words for that particular passage word (\ref{attended_p}). This is the new question-attended passage representation $v_t^P$. We further concatenate this with the original passage representation $u_t^P$ followed by a sigmoid gating (\ref{gated}) to represent how important this passage time step is for the encoding.

The penultimate stage of the attentive layer contains a bi-directional GRU (\ref{attentive_layer}) that acts a smoothing layer over the concatenation of original and question-attended passage representation.

We concatenate $c_T$ and $u_T^Q$  and perform a non-linear transformation of it before passing them to the decoder.
			\begin{align}
            c_T &= [\overleftarrow{c_T};\overrightarrow{c_T}]; h = \mathrm{tanh}(W_h [c_T;u_T^Q]) \label{encoder_output}
			\end{align}
            
Figure 2 shows a particular example of how relationship between question and passage is modeled.

\subsection{Decoding}

			\begin{table*}[h]
				\centering
				\begin{tabular}{c p{13cm}}
					\multicolumn{2}{l}{\textbf{}} \\
					& $\textbf{Q}$: what is a urethra \\
					& $\textbf{P}$: in anatomy , the urethra is a tube that connects the urinary bladder to the urinary meatus for the removal of fluids from the body . 1 infection of the urethra is urethritis , said to be more common in females than males . 2 urethritis is a common cause of dysuria ( pain when urinating ) . 3 related to urethritis is so called urethral syndrome . 4 passage of kidney stones through the urethra can be painful , which can lead to urethral strictures . \\
					& $\textbf{Model with coverage}$: is a tube that connects urinary bladder to the urinary meatus for the removal of fluids from the body ...... \\
                    & $\textbf{Model without coverage}$: the urethra a tube that connects urinary urinary bladder the the tube that connects the urinary bladder fluids the urinary body \\
				\end{tabular}
                \caption{Here are two types of repetitions we notice: consecutive identical (two times "urinary") words and duplicate phrases ("tube that connects the urinary bladder"). We show two results from two models -- with and without coverage}
				\label{repetition}
			\end{table*}
            
We use the GRU cell to compute our decoder states and the same scaled attention as in \ref{scaled_attention}.\\
The decoder decodes the information encoded by the query-aware passage encoder, aggregating information from various parts in the passage to produce the final answer. As we have seen in Figure 2, different regions of the passage can be weighted more if they are relevant to the query. In such cases, often we have seen from question-passage heatmaps, several areas in the passage are marked important. For eg. when the query is asking for some \textit{year}, all the words in the passage which denotes a year stands out more or less from other passage words (this issue is also highlighted in Table 1). It is now importantly the decoder's job to pick the correct year. The decoder is initialized with combination of both question representation encoded passage representation (from \ref{encoder_output})

\textbf{Making the RNN attention-aware}: We use the attended context concatenated with previous decoder state as input to the RNN to compute the current state. This makes the RNN aware about previous attentions. This differs from \citep{DBLP:conf/emnlp/LuongPM15}, where they attend after RNN state computation. We have observed that our difference causes a gain of at least 1 ROUGE-L point. The decoder state computation can be expressed as:
          \begin{align}
          h^-&=tanh(W_hh_i); s_{t-1}^-=tanh(W_ss_{t-1}) \nonumber \\
          e_i^t &= \frac{1}{\sqrt[]{D}} (W_h h_i^- W_s s_{t-1}^-)  \\
          a^t &= \text{softmax}(e^t) \nonumber \\
          h^*_t &= \sum\nolimits_i a_i^t h_i; s_{t-1}^*= \tanh(W_c[h_t^*; s_{t-1}]) \nonumber \\
          s_t &= GRU(s^*_{t-1}, y_{t-1}) \nonumber \\
          \end{align}
where $a_t$ is the attention distribution over the encoder states at $t^{th}$ time-step in the decoder and $s_t$ is the $t^{th}$ decoder hidden state. The attention methodology used here is identical to the one used in the attentive layer. Based on $a_t$, the decoder decides which encoder state to focus more on. $h_t^*$ is the attended context vector of the encoder at $t_{th}$ decoding time step. We use this context vector in concatenation with the decoder GRU's previous state $s_{t-1}$ to compute current state $s_t$. Finally, we compute the decoder output probability as $P_{vocab} = \text{softmax}(W_y s_t + b_y)$. This is a \textit{fully} abstractive approach.
\subsubsection{Copying from source}
However, to overcome the limitation of missing out entities from \textbf{P} (Table 1) or when dealing mostly extractive data, we apply pointer-generator networks to make it \textit{abstractive-cum-extractive}.With pointer-generator model, at each decoder time-step, we make a \textit{probabilistic} switch between whether to copy or simply choose from output distribution $P_{vocab}$. This is governed by $p_{gen}$ which is computed as (\citet{DBLP:journals/corr/SeeLM17}):
        \begin{align}
        p_{gen} = \sigma(w_h^T h^*_t + w_s^T s_t + b_\text{ptr})
        \end{align}
At each decoder time step, we use the decoder state $s_t$ and the context vector $h^*_t$, to decide whether to copy from the encoder words or to generate the highest probable word from the decoder output.

		\begin{align}
		P(w) &= p_{gen} p_{vocab}(w) + (1-p_{gen})\sum\nolimits_{i: w_i = w} a^t_i
		\end{align}

From a high-level point of view, the pointer-generator model, at each decoder step $t$, adds the attention probabilities of each encoder words from $a_t$ to their respective probabilities in the decoder output distribution $p_{vocab}$ at that decoder time step. A sigmoid over the affine transformations produce a probability. Higher $p_{gen}$ means the model will mostly choose a word from vocabulary while lower means higher chance of copying a source word from passage. $P(w)$ is the final probability distribution which is used for generating answer word and computing the training loss.
\subsubsection{Mitigating repetitions through coverage}
Readability is a major concern in almost all NLG tasks (Table 2). The copying mechanism helps to preserve information from the encoder side and prevent repetition of words which we have observed in our experiments. We realized that alleviating this problem will eventually lead to better readability of answers. Hence, we incorporated \textit{coverage mechanism}, which is pretty popular approach in machine translation, summarization, into our model. It is a mechanism in MT to avoid over-translation by keeping track of which source words are receiving attention too many times. Specifically, we used the attention distributions being computed at each decoder time step to maintain a coverage vector $cov_t$ which is basically cumulative sum of attention probabilities.
		\begin{align}
		cov_t &= \Sigma_{i=0}^{t-1} a_i \nonumber
		\end{align}
Hence, at each time step, the decoder has information about how much each encoder state has been attended until the previous step. We add an extra term $W_c cov_t$ to (7). This makes the standard attention mechanism has a knowledge about which states has been attended already enough in the past and thus manages to curb repetitions.

\subsection{Loss}
At each decoder time step, we use the negative log-likelihood of the correct word and finally try to minimize the total loss over all the decoder steps.
		\begin{align}
		loss &= - \frac{1}{T}\sum\nolimits_{t=0}^T \log P(w^*_t)
		\end{align} 
\begin{table}[t!]
\begin{center}
\begin{tabular}{|l|rl|}
\hline \bf Model & \bf  ROUGE-L & \bf Perplexity \\ \hline
Seq2Seq baseline & -/37.70 & -/- \\
gQA (+p-gen,+ cov.) & \textbf{74/59.5} & 4.35/4.62 \\
gQA (+p-gen) & 70/57.5 & 3.36/4.05 \\
gQA (w/o p-gen) & 45/42. & 4.3/4.48\\ 
\hline
\end{tabular}
\end{center}
\caption{The best performance on the MS-MARCO development data, given the correct passage. We report the perplexity alongside ROUGE-L score. For each numeric column, score follows the \textit{train/dev} format.}
				\label{marco_dev_result}
\end{table}
			\begin{figure}
				\begin{center}
					\includegraphics[width=3in]{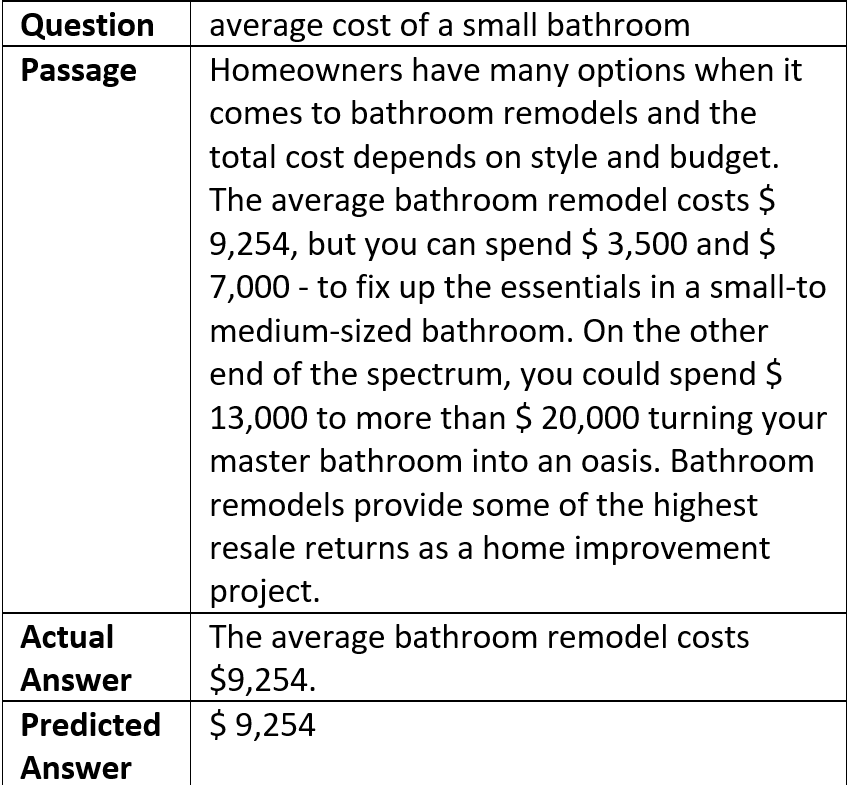}
				\end{center}
				\caption{Example result: core information retained}
				\label{res2}
			\end{figure} 
\section{Data and Experiments}
We conducted our experiments on MS-MARCO\footnote{Data is available for download at http://www.msmarco.org/dataset.aspx} data (\citet{DBLP:conf/nips/NguyenRSGTMD16}). In our experiments, to form the passage-answer pair from training data, we select the correct input passage for an answer by determing which passage has a sub-span with highest ROUGE-L score with the reference answer \citet{DBLP:journals/corr/TanWYLZ17}. We also verify that this ROUGE-L score is not less than 0.7. The filtered train data consists of 75000 examples. For evaluation, we take the correct passage for each query and generate answer from them. We relied on Stanford CoreNLP \citet{DBLP:conf/acl/ManningSBFBM14} to tokenize all texts. Also, We restricted the number of words in it to 30000 and lengths of P and A to 200 and 50 respectively. We use a batch of 50 examples for updating our model while training for roughly 15000 iterations. We use \textit{Glove} (\citet{DBLP:conf/emnlp/PenningtonSM14}) to represent words and keep them \textit{fixed}. We use hidden state dimension of 256 throughout the network. We use the Adadelta optimizer (\citet{DBLP:journals/corr/abs-1212-5701}), with \textit{epsilon}=1e-6, \textit{rho}=0.95 and initial learning rate=1, to minimize our loss. 
\section{Results and Analysis}
\begin{figure}
				\begin{center}
					\includegraphics[width=3in]{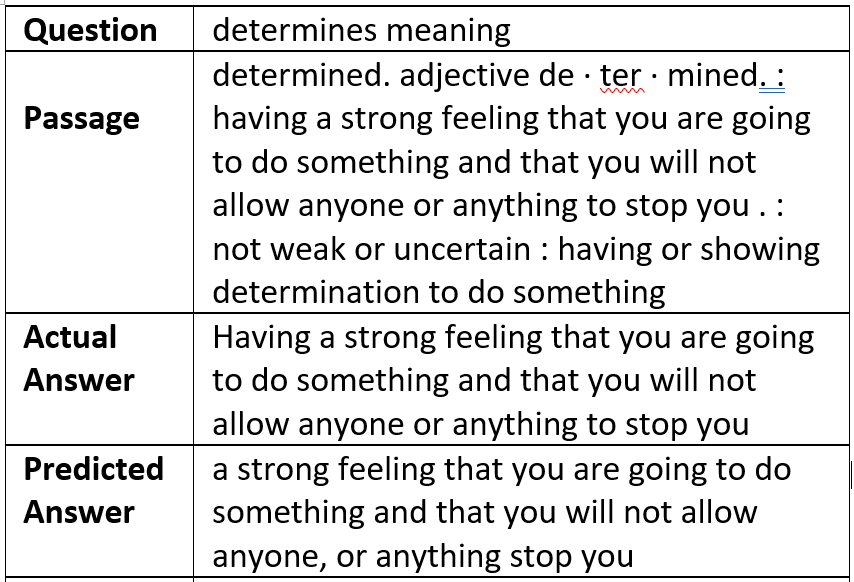}
				\end{center}
				\caption{Example result: almost same answer}
				\label{res}
			\end{figure}

We list our result on MS-MARCO data in Table \ref{marco_dev_result}. We do not consider negative passages and test the model only on correct passages for each query. We also include the results reported in Table 6 of \citet{DBLP:journals/corr/TanWYLZ17} on experimenting with basic sequence-to-sequence model with selected passage as the input. Results show that our model outperforms the generative baseline by a large margin. 
The learnt mean value of $p_{gen}$ is mostly 0.7 which tells us that the data is mostly extractive in nature and forces the model to copy mostly. We also tried to use self-attention but that didn't improve performance, most probably, because the decoder attention is already aggregating information from different parts of passage. Not only improvement in score, but also coverage reduced repetitions in multiple instances.
We provide some examples in Fig.\ref{res2} and Fig.\ref{res}.
\section{Conclusion}
We successfully build a generative model that can not only efficiently model relationship between question and passage but also can generate answers from the encoded relationship. We let the model decide to be in abstractive or extractive mode based on the nature of the data thus removing any dependency on any other external extractive feature. Moreover, we apply coverage in this QA task to mitigate frequent repetitions.

\bibliography{naaclhlt2018}
\bibliographystyle{acl_natbib}
\end{document}